\pdfoutput=1

\documentclass[11pt]{article}

\usepackage[final]{EMNLP2023}
\usepackage{times}
\usepackage{float}
\usepackage{hyperref}
\usepackage{latexsym}
\usepackage{multirow}

\usepackage[T1]{fontenc}

\usepackage[utf8]{inputenc}

\usepackage{microtype}
\usepackage{graphicx}
\usepackage{subfigure}
\usepackage{booktabs} 
\usepackage{amsmath}
\usepackage{amssymb}
\usepackage{mathtools}
\usepackage{amsthm}
\usepackage{adjustbox}
\usepackage{enumitem}

\usepackage{inconsolata}
\usepackage{pifont}
\usepackage{tikz}
\newcommand*\circled[1]{\tikz[baseline=(char.base)]{
            \node[shape=circle,draw,inner sep=0.4pt] (char) {#1};}}

\definecolor{mydarkblue}{rgb}{0,0.08,0.45}
\definecolor{myblue}{HTML}{3b75c9}
\definecolor{myred}{HTML}{E33222}
\definecolor{mygreen}{HTML}{438773}
\definecolor{mymaroon}{RGB}{142,27,19}
\definecolor{maroon}{HTML}{992000}
\definecolor{mycite}{cmyk}{0.55,1,0,0.15}
\definecolor{codeblue}{rgb}{0.25,0.5,0.5}
\definecolor{codekw}{rgb}{0.85, 0.18, 0.50}
\definecolor{codegreen}{rgb}{0,0.6,0}
\definecolor{codegray}{rgb}{0.5,0.5,0.5}
\definecolor{codepurple}{rgb}{0.58,0,0.82}
\definecolor{backcolour}{rgb}{0.95,0.95,0.92}
\definecolor{refcolor}{HTML}{9F363A}
\definecolor{dartgreen}{HTML}{00693e}

\hypersetup{
    colorlinks=true,
    citecolor=dartgreen,
    linkcolor=refcolor,
    urlcolor=dartgreen,
          }

\newif\ifshowcomments
\showcommentstrue 
\ifshowcomments
\fi

\usepackage{xspace}

\newcommand{\ourmethod}{\texttt{AlphaLoRA}\xspace}

\newcommand{\ALPHAHILL}{\texttt{PL\_Alpha\_Hill}\xspace}
\newcommand{\stablerank}{\texttt{Stable\_Rank}\xspace}
\newcommand{\alphahat}{\texttt{Alpha\_Hat}\xspace}

\title{\ourmethod: Assigning LoRA Experts Based on Layer Training Quality}

\author{
Peijun Qing$^{1}$ \quad Chongyang Gao$^{2}$ \quad Yefan Zhou$^{1}$ \quad Xingjian Diao$^{1}$ \\
{\bf Yaoqing Yang$^{1}$} \quad {\bf Soroush Vosoughi$^{1 \dagger}$} \\
$^{1}$Dartmouth College \quad $^{2}$Northwestern University \\
\texttt{\{peijun.qing.gr, yefan.zhou.gr, yaoqing.yang, soroush.vosoughi\}@dartmouth.edu} \\
\texttt{chongyanggao2026@u.northwestern.edu}
}

\begin{document}
\maketitle

\renewcommand{\thefootnote}{\fnsymbol{footnote}} 
\footnotetext[2]{Corresponding author.}  
\renewcommand{\thefootnote}{\arabic{footnote}}

\begin{abstract}
Parameter-efficient fine-tuning methods, such as Low-Rank Adaptation (LoRA), are known to enhance training efficiency in Large Language Models (LLMs). Due to the limited parameters of LoRA, recent studies seek to combine LoRA with Mixture-of-Experts (MoE) to boost performance across various tasks. However, inspired by the observed redundancy in traditional MoE structures, previous studies identify similar redundancy among LoRA experts within the MoE architecture, highlighting the necessity for non-uniform allocation of LoRA experts across different layers. In this paper, we leverage Heavy-Tailed Self-Regularization (HT-SR) Theory to design a fine-grained allocation strategy. Our analysis reveals that the number of experts per layer correlates with layer training quality, which exhibits significant variability across layers. Based on this, we introduce \ourmethod, a theoretically principled and training-free method for allocating LoRA experts to further mitigate redundancy. Experiments on three models across ten language processing and reasoning benchmarks demonstrate that \ourmethod achieves comparable or superior performance over all baselines. Our code is available at \href{https://github.com/morelife2017/alphalora}{https://github.com/morelife2017/alphalora}.
\end{abstract}

\section{Introduction}
LLMs have shown impressive performance on various NLP tasks \citep{brown2020language, chowdhery2022palm, touvron2023llama2, jiang2023mistral, jian-etal-2024-expedited, zhang-etal-2024-working-memory}. However, due to the increasing size of modern LLMs, significant computational resources are required for full fine-tuning. To address this issue, researchers are increasingly focusing on PEFT methods to reduce training costs, such as Adapter-tuning~\cite{houlsby2019parameter} and LoRA~\cite{hu2021lora}. Despite their training efficiency, the performance of PEFT methods in fine-tuning LLMs is still limited due to the small number of parameters \citep{xu2023parameter}.

To address the limitation, recent studies seek to combine LoRA and MoE by adding multiple LoRA modules \citep{liu2023moelora, gao2024higher,luo2024moelora}. The MoE structure in LLMs typically consists of multiple feed-forward "sub-networks", or "experts", that are trained to handle different types of inputs or tasks \citep{shazeer2017outrageously}. MoE structures are designed to dynamically activate only a subset of experts for each input, significantly scaling up the number of parameters, while incurring an affordable computational overhead. 

Existing LoRA-MoE methods integrate multiple LoRA modules into each sublayer of the transformer block and employ different strategies to assign experts to different tokens or tasks \citep{li2024mixlora, huang2023lorahub, zhang2023composing, yang2024moralmoeaugmentedlora, dou-etal-2024-loramoe, luo2024moelora, feng-etal-2024-mixture, liu2023moelora}. For example, at the task level, \citet{huang2023lorahub} and \citet{zhang2023composing} explore various composition strategies to combine multiple LoRA experts trained individually on different tasks. At the token level, both experts and input tokens interact with a routing network, resulting in the activation of experts based on the characteristics of the input tokens \citep{gao2024higher, dou2023art}. These methods demonstrate superior performance in single-task and multi-task learning compared to standard LoRA, with the number of experts per layer uniformly distributed, such as three per layer, as shown in Table~\ref{table:comparison_loramoe}.

However, recent studies \citep{chen2023sparse, zoph2022st} in MoE show that employing a large number of experts may be redundant due to representational collapse or learned routing policy overfitting.
Similarly, \citet{gao2024higher} investigate redundancy in parameter-efficient MoE. Unlike existing LoRA-MoE methods that uniformly allocate experts across all layers, they manually design four architectures with varying group-wise expert allocations. Specifically, they divide the 32 layers of the LLaMA-2 model \cite{touvron2023llama2} into 4 groups, with the first 8 layers constituting the first group, and so forth. They allocate a varying number of experts to each group (layers within the same group have the same number of experts). While their allocation strategy provides insights into overall architecture design, suggesting that higher layers need more LoRA experts, research on achieving more effective integration remains in its early stages.

To create a theoretically sound allocation strategy aimed at minimizing expert redundancy, our research draws inspiration from the Heavy-Tailed Self-Regularization (HT-SR) Theory~\citep{martin2019traditional, martin2020heavy, martin2021implicit, martin2021predicting}.
The HT-SR theory examines the properties of heavy-tailed (HT) structures observed in the Empirical Spectral Density (ESD) of weight matrices. The application of HT-SR to model selection and layer-wise adaptive training \citep{zhou2024temperature, yang2023test} showcases the theory's effectiveness in assessing both model and layer quality.

In this paper, we propose a fine-grained strategy for allocating layer-wise expert numbers, namely, \ourmethod. According to \citet{zhou2024temperature}, layers with more pronounced HT properties are generally well-trained. Following \citet{zhou2024temperature}, we measure the HT characteristics by fitting a power law (PL) distribution to the ESD and use the exponent as the metric to measure HT properties. We then use the Hill estimator~\citep{hill1975simple,zhou2024temperature} to calculate \ALPHAHILL. The core idea behind \ourmethod is to adaptively allocate LoRA experts based on layer training quality, which is indicated by \ALPHAHILL values, thus reducing redundancy in a more theoretically-principled manner. The contributions of our work are summarized as follows:

\begin{itemize}[leftmargin=*]
    \item We are the first to interpret the correlation between layer-wise training quality and LoRA expert number through the lens of HT-SR theory. Empirical results on three widely-used language models suggest that well-trained layers need fewer LoRA experts. 
    \item We propose a fine-grained allocation strategy, \ourmethod, for allocating layer-wise expert numbers. Inspired by HT-SR theory, this method is theoretically grounded and training-free. \ourmethod generally outperforms MoLA-$\triangledown$, the current state-of-the-art non-uniform expert allocation method, across three models and ten NLP datasets. Notably, \ourmethod with 80 experts surpasses MoLA-$\triangledown$(2468) with 160 experts, achieving comparable performance with 50\% fewer parameters.
    \item We compare several layer-wise weight matrix metrics from HT-SR theory for evaluating layer training quality and allocating expert numbers. The relative performance of these metrics reveals that the \ALPHAHILL metric outperforms others, corroborating the findings of \cite{zhou2024temperature} that the \ALPHAHILL metric is better for assessing layer training quality. This further demonstrates the correlation between layer expert number and layer training quality.

\end{itemize}

\begin{table}[!ht]
    \centering    
    \begin{tabular}{l|l}
        \toprule        
        \textbf{Granularity} & \textbf{Method} \\
        \midrule
        
        \multirow{8}{*}{\textbf{Uniform}}& MixLoRA\cite{li2024mixlora} \\ 
        & MoRAL \cite{yang2024moralmoeaugmentedlora} \\ 
        & LoRAMoE \cite{dou-etal-2024-loramoe} \\ 
        & MoELoRA \cite{luo2024moelora} \\ 
        & MoA \cite{feng-etal-2024-mixture} \\ 
        & MOELoRA \cite{liu2023moelora} \\ 
        \midrule
        \textbf{Group-wise}& MoLA \cite{gao2024higher} \\ 
        \midrule
        \textbf{Layer-wise}& \textbf{\ourmethod} \\
        \bottomrule    
    \end{tabular}

    \caption{Comparison of allocation strategy between different LoRA-MoE methods.}
    \label{table:comparison_loramoe}
\end{table}

\begin{figure*}[!ht]
\begin{center}
\centerline{\includegraphics[width=2\columnwidth]{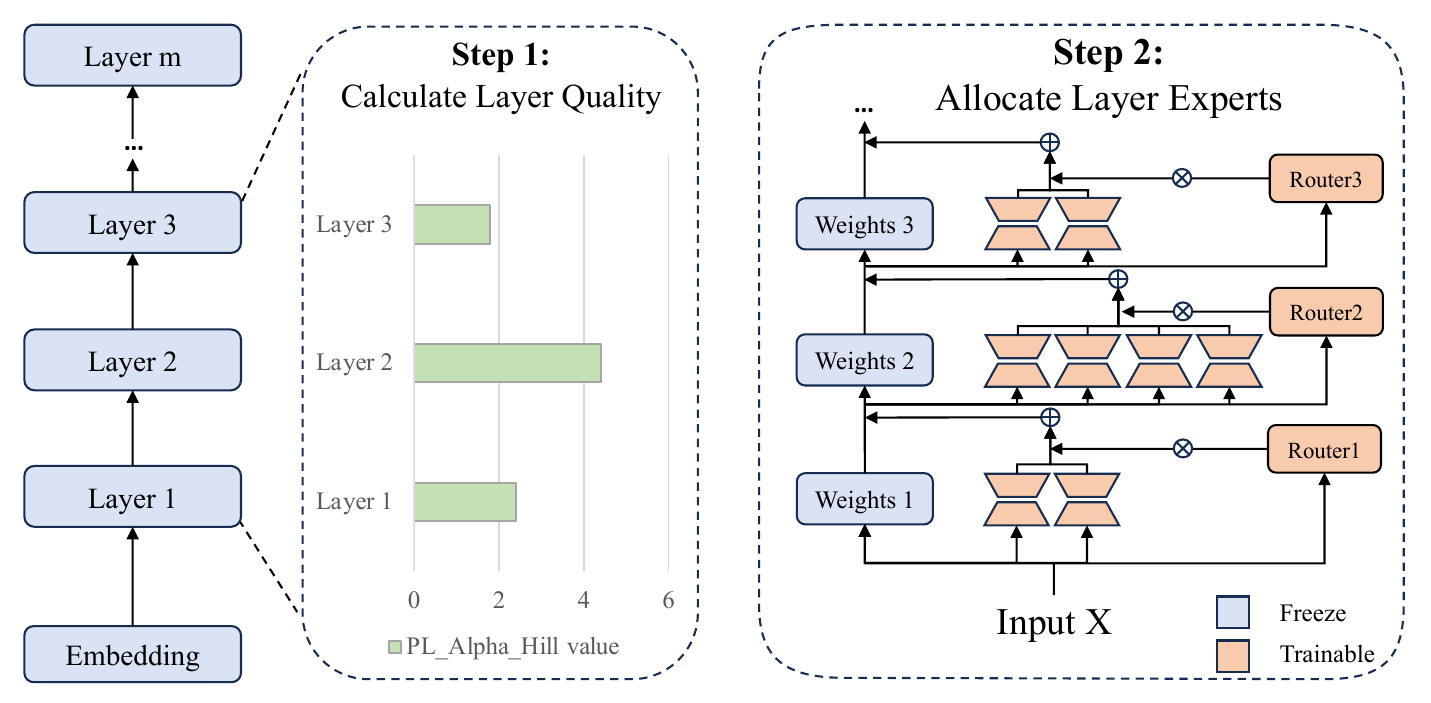}}
\caption{The overview of \ourmethod. For a transformer-based model with $m$ layers, \ourmethod involves two steps: (Step 1) Conducting ESD analysis on each layer and applying PL fitting to obtain the layer-wise \ALPHAHILL value. (Step 2) Converting the layer-wise \ALPHAHILL value into the number of experts using a mapping function, followed by initializing the experts for each layer. For instance, Weights 1 represents all the weight matrices (such as attention weight matrix and projection weight matrix) in layer 1.}
\label{fig1:alpha}
\vspace{-5mm}
\end{center}
\end{figure*}

\section{Preliminary}

\subsection{LoRA-MoE Architecture}
\label{sec:MoLA_architecture}

The LoRA-MoE architecture creates multiple LoRA experts for each layer in a pre-trained LLM. The “LoRA expert” used in this work refers to the vanilla LoRA block \cite{hu2021lora}. For a pre-trained weight matrix $\mathbf{W_0} \in \mathbb{R}^{m\times n}$ ($n < m$), LoRA creates two low-rank trainable matrices $\mathbf{A}$ and $\mathbf{B}$, where $\mathbf{A} \in \mathbb{R}^{m\times r}$, $\mathbf{B} \in \mathbb{R}^{r\times n}$, and $r \ll \min(m, n)$. Thus, the dimension of $\mathbf{A}\mathbf{B}\mathbf{x}$ equals the dimension of $\mathbf{W_{0}x}$ for the input $\mathbf{x}$. During training, $\mathbf{W_0}$ is frozen while $\mathbf{A}$ and $\mathbf{B}$ receive gradient updates. The output $\mathbf{h}$ is expressed as follows:
\begin{equation}
\mathbf{h} = \mathbf{W}_{0}\mathbf{x} + \Delta \mathbf{Wx} = \mathbf{W}_{0}\mathbf{x} + \mathbf{ABx}.
\label{eq2:lora}
\end{equation}
Each LoRA-MoE layer contains $N$ LoRA experts, which is denoted as $\{\sum_{i=1}^{N}\}$. The forward process of the layer is expressed as:
\begin{equation}
\mathbf{o} = \mathbf{W}_0 \mathbf{x} + \Delta \mathbf{W x} = \mathbf{W}_0 \mathbf{x}+ \sum_{i=1}^{N} \mathbf{G(x)}_i \mathbf{E}_i(\mathbf{x}),
\end{equation}
where \( \mathbf{E}_i(\mathbf{x}) \) and \( \mathbf{G}(\mathbf{x}) = \text{Softmax}(\mathbf{x} \mathbf{W}_g) \) represent the \( i \)-th expert and the router in the LoRA-MoE layer, respectively.
The \( \mathbf{W}_g \) is the trainable parameter matrix of the route network used to allocate input \(x\) to different experts. The router enables experts to develop varied capabilities and efficiently handle various types of tasks and inputs.

\subsection{ESD Shape Metric}
\paragraph{Empirical Spectral Density (ESD).} ESD represents the distribution of eigenvalues of a matrix, often used to understand properties of large random matrices that arise in various applications such as neural networks~\citep{martin2021implicit}. Let $\mathbf{A} = $\(\mathbf{W}_{0}^{\top}\mathbf{W}_0 \in \mathbb{R}^{n \times n}\) be a symmetric matrix with eigenvalues $\{\lambda_i^{\mathbf{A}}\}_{i=1}^n$. The empirical spectral density (ESD) of \(\mathbf{A}\) is defined as the probability measure:
\begin{equation}
\rho(\lambda; \mathbf{A}) = \frac{1}{n} \sum_{i=1}^n \delta \left( \lambda - \lambda_i^{\mathbf{A}} \right),
\end{equation}
where \(\delta\) is the Dirac delta function. The ESD represents the distribution of the eigenvalues of \(\mathbf{A}\).

\paragraph{Heavy-Tailed Self-Regularization (HT-SR) Theory.}
Drawing from Random Matrix Theory \citep{tao2023topics,bai2010spectral,couillet2022random}, HT-SR theory relies on the empirical observation that well-trained models usually display strong correlations, leading to HT structures in the ESD of each layer \citep{martin2021predicting,yang2023test,zhou2024temperature}. Derived from HT-SR theory, \emph{shape metrics} are analytical methods used to characterize the HT properties of ESDs in neural networks, which correlate with the shapes emerging in their ESDs. In this work, we mainly study three shape metrics: \ALPHAHILL, \texttt{Alpha\_Hat}, and \texttt{Stable\_Rank}. 
Inspired by previous work on estimating model quality \citep{zhou2024temperature, yang2023test}, we apply \ALPHAHILL, which is proved to be the most effective, as the main metric to evaluate layer quality. The definition of \ALPHAHILL is explained in \S\ref{sec:alpha_cal}. Other definitions can be found in the Appendix~\ref{appen:def_metric}.

\vspace{-3mm}
\section{Method}

In this section, we introduce \ourmethod, an expert allocation strategy based on the \ALPHAHILL metric. Given a transformer-based model, we first compute the \ALPHAHILL metric value for each layer. \S\ref{sec:alpha_cal} elaborates on the calculation of layer-wise \ALPHAHILL metric. \S\ref{sec:mapping_alpha} gives the process of mapping the layer-wise \ALPHAHILL value to expert number. The overview of our method is illustrated in Figure~\ref{fig1:alpha}.

\subsection{Analyzing Layer Training Quality}
\label{sec:alpha_cal}

\begin{figure}[!ht]
\begin{center}
\centerline{\includegraphics[width=1.05\columnwidth]{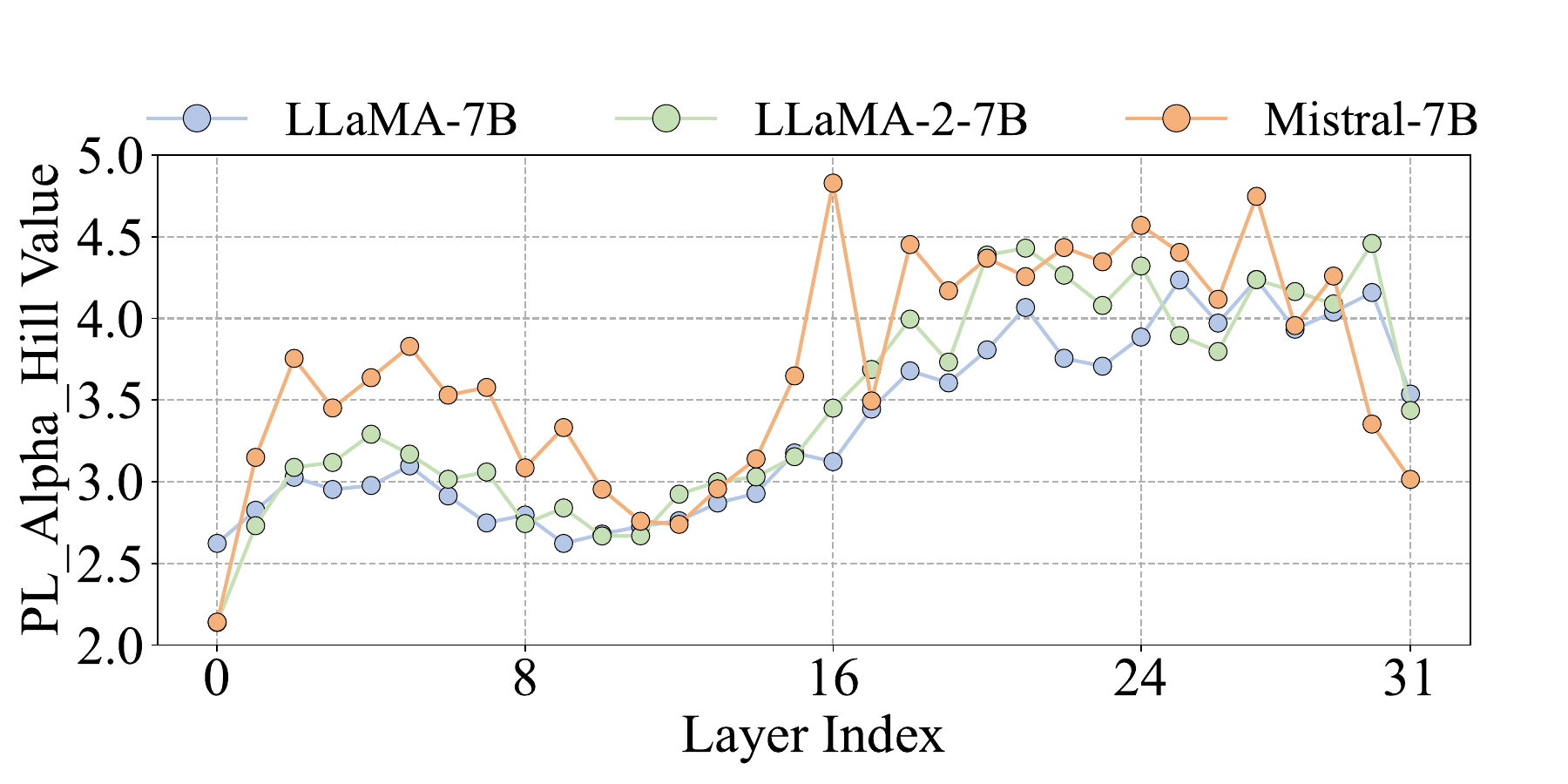}}
\caption{Illustration of the \ALPHAHILL values for each layer across three different models.}
\label{fig:2s}
\end{center}
\end{figure}

\begin{figure*}[!ht]
\begin{center}
\centerline{\includegraphics[width=2.1\columnwidth]{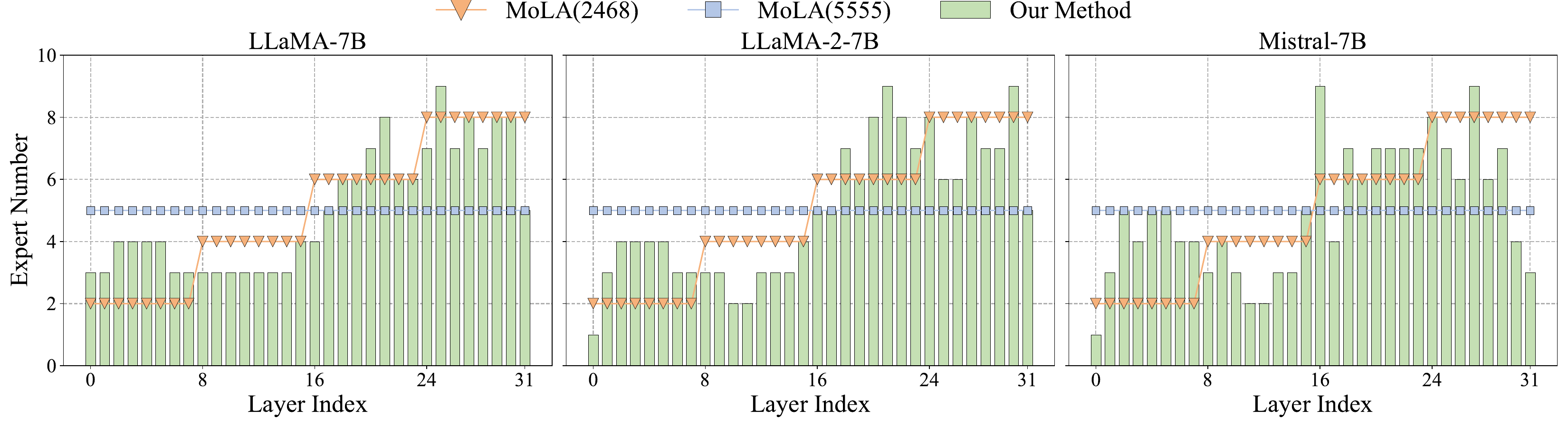}}
\caption{Comparing layer expert number assigned by \ourmethod and MoLA. MoLA(2468) allocates \textbf{2} experts to each layer for the first 8 layers, \textbf{4} experts to each layer for 9-16 layers, \textbf{6} experts to each layer for 17-24 layers, and \textbf{8} experts to each layer for the last 8 layers, which is denoted as \textbf{2468}. MoLA(5555) assigns a uniform \textbf{5} experts to each layer. The total number of experts is set at 160.}
\label{fig3_expert_allocation} 
\vspace{-5mm}
\end{center}
\end{figure*}

\ourmethod measures the layer training quality based on the HT characteristic of the layer ESDs, which is quantified by the HT metric \ALPHAHILL. For the $i$-th weight matrix in each layer from a transformer-based model, we first calculate the eigenvalues of its correlation matrix $\mathbf{X}_i = \mathbf{W}_i^T \mathbf{W}_i$ and ESD $\rho$. We then fit a power law (PL) distribution $p$ to the HT part of the ESD, taking values within an interval $(\lambda_{\min}, \lambda_{\max})$, which is defined as:
\begin{equation}
p(\lambda) \propto \lambda^{-\alpha}, \quad \lambda_{\text{min}} < \lambda < \lambda_{\text{max}}.
\end{equation}
We refer to its exponent $\alpha$ as \texttt{PL\_Alpha}, where a lower value means more Heavy-tailed.
We then use the Hill estimator~\citep{hill1975simple} to calculate \ALPHAHILL as the following:
\begin{equation}
\texttt{PL\_Alpha\_Hill} = 1 + \frac{k}{\left( \sum_{i=1}^{k} \ln \frac{\lambda_{n-i+1}}{\lambda_{n-k}} \right)},
\end{equation}
where $k$ is a parameter for controlling the lower eigenvalue threshold $\lambda_{\min}$ for PL estimation. We apply the Fix-finger method \cite{yang2023test} to select the $k$, which keeps $\lambda_{\min}$ at the peak of the ESD. We calculate \ALPHAHILL for each weight matrix within a layer separately and compute the average to represent the layer value. Figure~\ref{fig:2s} shows the \ALPHAHILL values for three popular language models. The metric values show non-uniform distributions across layers, indicating varying training quality between layers. 

Applications of HT-SR theory \cite{zhou2024temperature} indicate assigning larger learning rates to under-trained layers helps these layers capture more correlations (or features) from the data \cite{wang2024spectral}. This implies that under-trained layers, which have captured fewer features, may need more LoRA experts to acquire additional features during fine-tuning.

\subsection{Allocating Expert Based on Layer Quality}
\label{sec:mapping_alpha}
Given a Transformer model with \( m \) layers, we use \( s_i \) to denote the number of experts in layer \( i \). We first calculate the \ALPHAHILL metric value for each layer using the algorithm in \S\ref{sec:alpha_cal} to obtain a sequence of metric values \( \mathcal{V} = [v_1, v_2, \ldots, v_m] \).

The number of experts assigned to each layer is determined using a mapping function \(\psi: \mathbb{R}^m \rightarrow \mathbb{R}^m\). This function converts a sequence of metric values \(\mathcal{V} = [v_1, v_2, \ldots, v_m]\) into the corresponding expert numbers \(\mathcal{S} = [s_1, s_2, \ldots, s_m]\), represented as:
\begin{equation}
s_i = \left\lfloor \left(\frac{v_i^{\beta}}{\sum_{i=1}^{m} v_i^{\beta}}\right) \times T \right\rfloor
,\end{equation}
where \( \lfloor \cdot \rfloor \) denotes rounding to the nearest integer, and \(\beta\) is an exponent parameter that controls the standard deviation of the sequence \(\mathcal{S}\). The values in \(\mathcal{V}\) are normalized by dividing by their sum, ensuring that the total allocation across all layers is proportional to their relative importance. We introduce a target sum parameter \( T \) such that \( \sum_{i=1}^{m} s_i = T \). Multiplying by \(T\) scales this proportional allocation to the desired number of experts, and finally, rounding to the nearest integer ensures discrete expert numbers for each layer.

After rounding, if the total number of experts across all layers doesn't equal the target number \( T \), we iteratively increment or decrement the integer value as follows:
\begin{equation}
\begin{cases}
   \text{if} \ \sum_{i=1}^{m} s_i < T, & s_{\arg\min (v_i^e - s_i)} + 1 \\
   \text{if} \ \sum_{i=1}^{m} s_i > T, & s_{\arg\max (v_i^e - s_i)} - 1
\end{cases}
,\end{equation}
where we identify a particular index in $\mathcal{S}$ that has the minimum (or maximum) difference between the scaled value and the present integer value, indicating the specific integer that needs to be incremented (or decremented) when the current sum is below (or above) the target sum.
After getting the number of expert \( \mathcal{S} = [s_1, s_2, \ldots, s_m] \) for each layer, we allocate $s_i$ experts for layer $i$ in a transformer-based model with $m$ layers, resulting in $\sum_{i=1}^{m} s_i = T$ experts in total. Given a pre-trained weight matrix $\mathbf{W}_0^{i,t} \in \mathbb{R}^{m \times n }$ from module $t$ in layer $i$, we create $s_i$ pairs of low-rank matrices $\{\mathbf{A}^{i,t}_j, \mathbf{B}^{i,t}_j\}_{j=1}^{s_i}$. Each matrix $\mathbf{A}_j^{i,t}$ is initialized from a random Gaussian distribution, while $\mathbf{B}_j^{i,t}$ is set to zero, where $\mathbf{A}_j^{i,t} \in \mathbb{R}^{m \times r}$, $\mathbf{B}_j^{i,t} \in \mathbb{R}^{r \times n}$, and $r \ll \min(m, n)$ \cite{hu2021lora}.

A router $\mathbf{S}_j^{i,t}$ with a trainable weight matrix $\mathbf{W}_r^{i,t} \in \mathbb{R}^{n \times N_j}$ assigns experts for input $\mathbf{x}$. Following MoLA \citep{gao2024higher}, we use the top-$K$ strategy for computation and apply a load balancing loss at each layer \cite{zoph2022st}. This process is mathematically represented as follows:
\begin{equation}
\mathbf{S}_j^{i,t}(\mathbf{x}) = \frac{\text{TopK}(\text{Softmax}(\mathbf{W}_r^{i,t} \mathbf{x}), K)_j}{\sum_{j=1}^{K}\text{TopK}(\text{Softmax}(\mathbf{W}_r^{i,t} \mathbf{x}), K)_j}
\label{eq:router}
,\end{equation}
\begin{equation}
\mathbf{h}^{i,t} = \mathbf{W}_0^{i,t} \mathbf{x} + \sum_{j=1}^{K} \mathbf{S}_j^{i,t}(\mathbf{x}) \mathbf{A}_j^{i,t} \mathbf{B}_j^{i,t} \mathbf{x}
\label{eq:}
.\end{equation}
The output $\mathbf{h}^{i,t}$ is obtained by adding the transformation of $\mathbf{x}$ via the pre-trained weight matrix $\mathbf{W}_0^{i,t}$ to the aggregated low-rank transformations from the top $K$ experts, where each expert's contribution is modulated by its corresponding assignment probability $\mathbf{S}_j^{i,t}(\mathbf{x})$.

\section{Experiment}
We design two experimental settings to examine the performance of \ourmethod, including direct fine-tuning and instruction-tuning$\rightarrow$zero-shot evaluation. \S\ref{sec4.1: expertallocation} compares the expert allocation between \ourmethod and variants of MoLA \cite{gao2024higher}. \S\ref{sec4.2: main_results} presents the results for two experimental settings. Implementation details can be found in Appendix~\ref{appen: implement}.

\paragraph{Models and datasets.}
To demonstrate the effectiveness of our approach, we conduct evaluations on three LLMs: LLaMA-7B~\citep{touvron2023llama}, LLaMA-2-7B~\citep{touvron2023llama2} and Mistral-7B-v0.1~\citep{jiang2023mistral}. We evaluate both NLP tasks and reasoning tasks. For the first setting, following \citet{gao2024higher}, we assess the performance on three GLUE datasets and three commonsense reasoning datasets: (1) Microsoft's Research Paraphrase Corpus (MRPC) \citep{mrpc}, (2) Recognizing Textual Entailment (RTE) dataset \citep{glue}, (3) Corpus of Linguistic Acceptability (COLA) \citep{glue}, (4) ScienceQA \citep{scienceqa}, (5) CommonsenseQA \citep{commonsenseqa}, and (6) OpenbookQA \citep{openbookqa}. For the second setting, we evaluate arithmetic reasoning on four zero-shot benchmarks: (1) AddSub \citep{addsub}, (2) MultiArith \citep{multiarith}, (3) SVAMP \citep{patel2021svamp}, (4) GSM8K \citep{cobbe2021gsm8k}. The detailed description of each dataset is shown in Appendix~\ref{appen:dataset}.

\begin{table*}[!ht]
    \centering
    \setlength{\tabcolsep}{3.5pt} 
        \begin{tabular}    {l | l | c  c  c|  c  c  c | c }
        \toprule        
        Models & Strategy &  MRPC & COLA & RTE &  S.QA & C.QA & O.QA & Average \\
        \midrule
        \multirow{4}{*}{\textbf{LLaMA}}
        &\textbf{MoLA-$\square$ (8888)} & 82.55 &  84.37& 84.47 & 90.82 & 76.82 & 76.60 & 82.61 \\
        \cmidrule(lr){2-9}
        &\textbf{MoLA-$\square$ (5555)} & 82.43 & 84.18 & 83.03 & 90.28 & 75.10 & 76.00 & 81.84 \\
        &\textbf{MoLA-$\triangledown$ (2468)} & 83.36 & 84.64 & 84.83 & 90.10  & 75.42 & \textbf{78.60} & 82.83 \\
        &\textbf{\ourmethod} & \textbf{85.19} & \textbf{85.42} & \textbf{85.19} & \textbf{90.37} & \textbf{76.49} & 78.20 & \textbf{83.48} \\
        \midrule

        \multirow{4}{*}{\textbf{LLaMA-2}}
        &\textbf{MoLA-$\square$ (8888)} &84.70 & 85.81 & 88.45 & 91.91 & 77.89 & 82.80 & 85.26 \\
        \cmidrule(lr){2-9}
        &\textbf{MoLA-$\square$ (5555)} & 84.17 & 86.19 & 84.83 & 92.08 & 77.55 & 80.00 & 84.14 \\
        &\textbf{MoLA-$\triangledown$ (2468)} & 83.48 & \textbf{86.87} & 86.28 & 92.36 & \textbf{78.95} & 79.60 & 84.59 \\
        &\textbf{\ourmethod} & \textbf{84.23} & 86.67 & \textbf{87.36} & \textbf{92.71}  & 78.05 & \textbf{80.80} & \textbf{84.97} \\
        \midrule
        \multirow{4}{*}{\textbf{Mistral}}
        &\textbf{MoLA-$\square$ (8888)} & 86.43 & 87.24 &  89.53 & 94.91 & 82.96 & 88.60 & 88.28 \\
        \cmidrule(lr){2-9}
        &\textbf{MoLA-$\square$ (5555)} & 85.73 & 87.34 & 88.44 & 94.46 & 81.90 & 88.00 & 87.65 \\
        &\textbf{MoLA-$\triangledown$ (2468)} & 86.95 & 87.44 & 88.80 & \textbf{95.14}  & 83.37 & 88.20 & 88.32 \\
        &\textbf{\ourmethod} & \textbf{87.13} & \textbf{87.91} & \textbf{91.70} & 95.00  & \textbf{84.00} & \textbf{89.20} & \textbf{89.16} \\
        
        \bottomrule    
        \end{tabular}
        \caption{Accuracy comparison with different methods on direct fine-tuning (S.QA, C.QA, O.QA denote ScienceQA, CommonsenseQA, and OpenbookQA respectively). The total number of experts for MoLA-$\square$ (8888) is 256, while the other variants are 160. \ourmethod outperforms other variants or baselines and even achieves competitive or superior performance with MoLA-$\square$ (8888), with nearly 40\% fewer parameters.}
    \label{table1:finetuning}
\end{table*}

\paragraph{Baselines.}
We compare our method with MoLA, which involves the manual design of 4 different allocation strategies, with MoLA-$\triangledown$(2468) as the state-of-the-art method. Specifically, take LLaMA-2~\citep{touvron2023llama2} which contains 32 layers, as an example. MoLA-$\triangledown$(2468) allocates \textbf{2} experts to each layer for the first 8 layers, \textbf{4} experts to each layer for 9-16 layers, \textbf{6} experts to each layer for 17-24 layers and \textbf{ 8} experts to each layer for the last 8 layers, which is denoted as \textbf{2468}. Thus, the overall structure forms a $\triangledown$ shape. Following similar notation, MoLA-$\square$(5555) employs an allocation strategy of \textbf{5555}, uniformly assigning 5 experts to each layer. MoLA-$\square$(8888) uniformly assigns 8 experts to each layer. To ensure a fair comparison, we introduce a target sum parameter $T$ to control the total number of experts for \ourmethod, thereby equalizing the number of trainable parameters between \ourmethod and MoLA. The number of trainable parameters is 105,635,840, which is 1.5\% of the trainable parameters in the pre-trained base model. \looseness-1

\subsection{Analysis of Expert Allocation}
\label{sec4.1: expertallocation}
In Figure \ref{fig3_expert_allocation}, we present the allocation of experts across all layers of three LLMs: LLaMA-7B~\citep{touvron2023llama}, LLaMA-2-7B~\citep{touvron2023llama2}, and Mistral-7B-v0.1~\citep{jiang2023mistral}. Our allocation indicates that the middle layers are generally better trained than the higher and lower layers, suggesting that they should be assigned fewer LoRA experts. In contrast, the higher layers are less well-trained and require more experts. Our analysis reveals that the overall allocation of the three models forms a loosely "M" shape, a pattern not previously explored by \citet{gao2024higher}. Additionally, we note that Mistral-7B-v0.1 requires more experts in the lower layers compared to the LLaMA models, highlighting the need for a model-specific allocation strategy.

\begin{table*}[!ht]
    \centering

        \begin{tabular}    {l|l| c  c  c  c | c }
        \toprule        
        Models & Strategy &  GSM8K & SVAMP & AddSub & MultiArith & Average \\
        \midrule

        
        \multirow{3}{*}{\textbf{LLaMA}}& \textbf{MoLA-$\square$ (5555)} & 44.04 & 52.00 & 38.89 & \textbf{88.16} & 55.77 \\
        &\textbf{MoLA-$\triangledown$ (2468)} & 43.59 & 52.80 & 40.50 & 84.33 & 55.31 \\
        &\textbf{\ourmethod} & \textbf{45.03} & \textbf{53.60} & \textbf{42.27} & 86.66 & \textbf{56.89} \\
        \midrule
        
        \multirow{3}{*}{\textbf{LLaMA-2}}&\textbf{MoLA-$\square$ (5555)} & 49.50 & \textbf{57.10} & 47.08 & 87.00 & 60.17 \\
        &\textbf{MoLA-$\triangledown$ (2468)} & 50.11 & 56.40 & \textbf{48.86} & 87.66 & 60.76 \\
        &\textbf{\ourmethod} & \textbf{50.41} & 57.00 & 48.60 & \textbf{91.33} & \textbf{61.84} \\
        \midrule
        
        \multirow{3}{*}{\textbf{Mistral}} &\textbf{MoLA-$\square$ (5555)} & 69.37 & 73.60 & 56.45 & 96.00 & 73.86 \\
        &\textbf{MoLA-$\triangledown$ (2468)} & 67.20 & 76.50 & 59.24 & 97.00 & 74.99 \\
        &\textbf{\ourmethod} & \textbf{68.30} & \textbf{77.20} & \textbf{59.49} & \textbf{97.33} & \textbf{75.58} \\

        \bottomrule    
        \end{tabular}
        \caption{Accuracy comparison with different methods with same total experts number on zero-shot tasks evaluation. Each method is trained on the MetaMathQA dataset \cite{yu2023metamath} and evaluated on four mathematical datasets.}
        \vspace{-2mm}
    \label{table2:math}
    
\end{table*}
\subsection{Main Result}
\label{sec4.2: main_results}

\paragraph{Direct fine-tuning.} The first experimental setup adheres to the evaluation protocol detailed in MoLA~\cite{gao2024higher}. We perform direct instruction fine-tuning \citep{wei2021finetuned, sanh2022multitask} for various allocation strategies on six NLP datasets, assessing performance on their respective test sets. \\
\textbf{\circled{1} \ourmethod enhances efficiency in LoRA-MoE experts.} In comparison to MoLA-$\square$ (8888), which utilizes 256 experts, \ourmethod shows superior performance on LLaMA and Mistral models and achieves similar results on LLaMA-2, while using only 160 experts (62.5\% of the parameters compared to MoLA-$\square$ (8888)). For instance, \ourmethod surpasses MoLA-$\square$ (8888) by 2.17\% on the RTE dataset for the Mistral model. The results indicate that a uniform allocation strategy leads to redundancy across various models. \ourmethod ensures efficient allocation, enabling the experts to capture more knowledge.\\
\textbf{\circled{2} \ourmethod outperforms other baseline allocation methods.} Given an equal total number of experts, \ourmethod, based on layer training quality, consistently outperforms MoLA-$\square$ (5555) across three models, with performance improvements of 1.64\%, 0.83\%, and 1.51\%, respectively. Furthermore, \ourmethod surpasses MoLA-$\triangledown$ (2468) by 0.65\%, 0.38\%, and 0.84\%, respectively. This highlights the superiority of our adaptive layer-wise allocation strategy.\\
\textbf{\circled{3} Correlation between number of experts and layer training quality.} As shown in Figure~\ref{fig3_expert_allocation}, the overall allocation of \ourmethod is more similar to MoLA-$\triangledown$ (2468) compared to the difference with MoLA-$\square$ (5555), resulting in a relatively smaller performance improvement. This underscores the correlation between the number of layer experts and training quality, suggesting that well-trained layers require fewer LoRA experts.

\paragraph{Zero-shot tasks.} In the second setting, we perform instruction-tuning on the MetaMathQA dataset \cite{yu2023metamath} and evaluate on four zero-shot benchmarks: GSM8K \cite{cobbe2021gsm8k}, SVAMP \cite{patel2021svamp}, AddSub \cite{addsub}, and MultiArith \cite{multiarith}. This setting evaluates the transfer learning capabilities of different allocation strategies. As shown in Table~\ref{table2:math}, \ourmethod outperforms both the state-of-the-art allocation strategy MoLA-$\triangledown$ (2468) and the uniform allocation MoLA-$\square$ (5555). For example, \ourmethod exceeds MoLA-$\triangledown$ (2468) by an average of 1.58\% across the four benchmarks. The results corroborate our earlier findings from direct fine-tuning experiments, demonstrating that a fine-grained allocation strategy like \ourmethod enables models to capture more knowledge and reduce redundancy among experts, thereby enhancing overall performance.

\subsection{Layer Quality Metrics for Expert Allocation}
\label{sec4.3: shapemetric}
\begin{figure}[!ht]
\begin{center}
\centerline{\includegraphics[width=1.1\columnwidth]{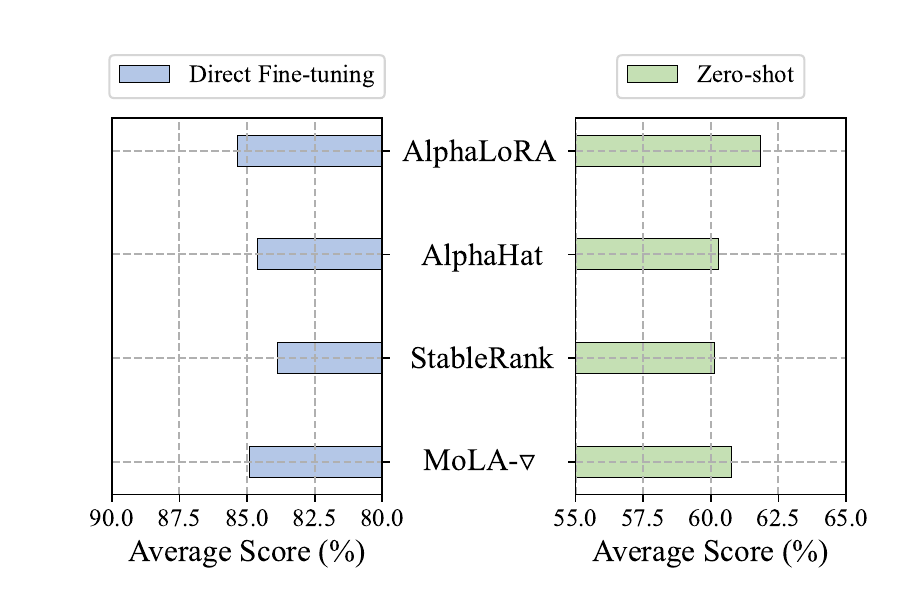}}
\caption{Comparison with different shape metric from HT-SR theory on both direct fine-tuning and zero-shot setting.}
\label{fig: shapemetric}
\end{center}
\end{figure}

In this study, we evaluate several shape metrics in HT-SR theory for measuring layer training quality. The definition of other metrics can be found in Appendix~\ref{appen:def_metric}.  Experiments are conducted on LLaMA-2 \cite{touvron2023llama2} under the same parameter setting for each task. Figure~\ref{fig: shapemetric} shows that \ALPHAHILL outperforms other shape metrics on both direct fine-tuning and zero-shot setting, aligning with the findings of \cite{zhou2024temperature} that the \ALPHAHILL metric is better for assessing layer training quality. This further demonstrates that layer expert number correlates with layer training quality. The detailed results are shown in Table~\ref{table:results1}, ~\ref{table:results2} and~\ref{table:shape_metrics_zeroshot}, Appendix~\ref{appen: detail_resutls_shapem}.

\subsection{Consistent Superiority with Varying Number of Experts}
\label{sec4.4: expertnumber}
\begin{figure}[!ht]
\begin{center}
\centerline{\includegraphics[width=1\columnwidth]{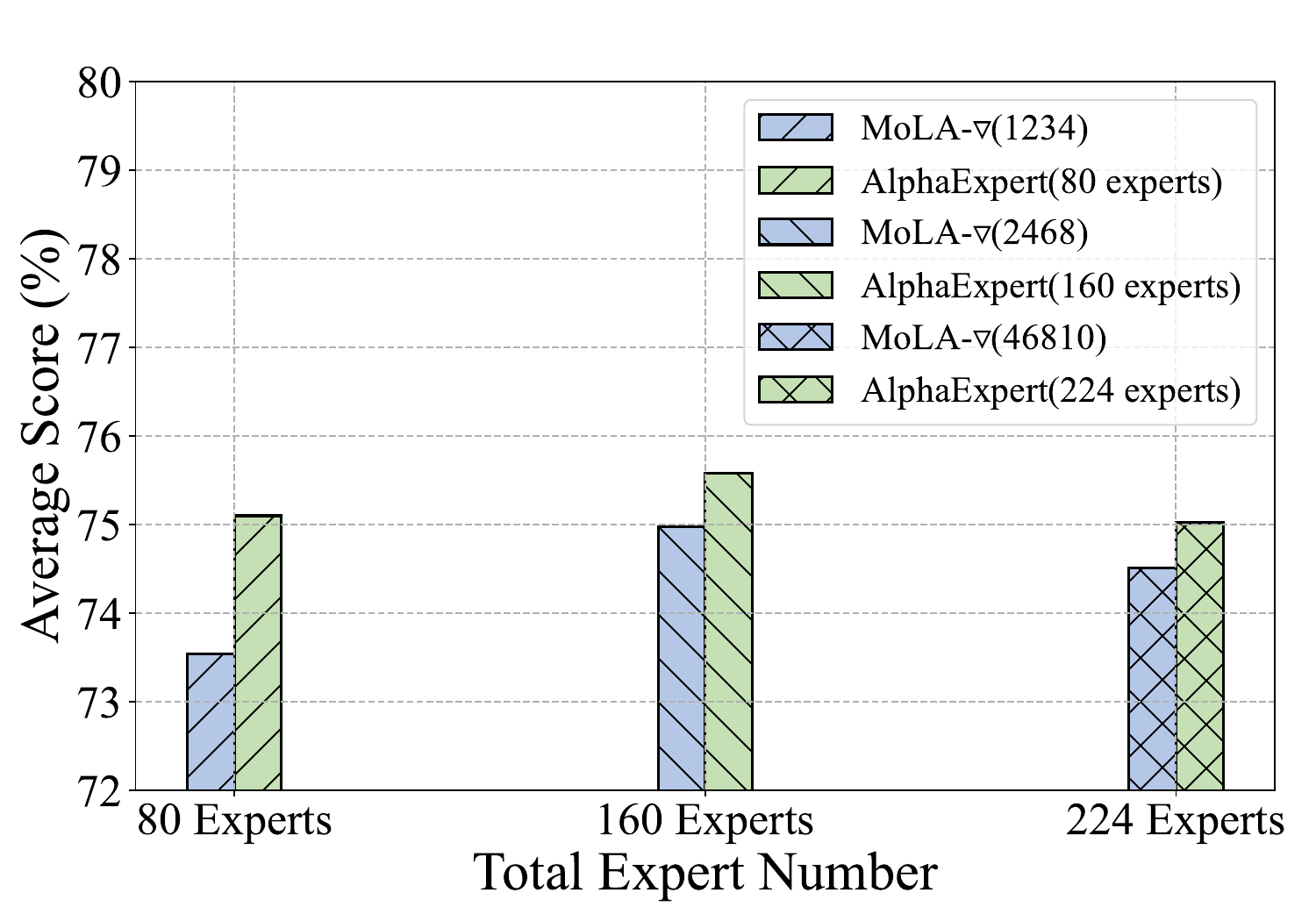}}
\caption{Comparison between \ourmethod and MoLA-$\triangledown$ across three configurations with varying total number of experts $T$, specifically 80, 160, and 224 experts. }
\label{fig:expert_number}
\end{center}
\end{figure}

In this study, we compare \ourmethod with MoLA-$\triangledown$ across three configurations on Mistral-7B model, each with a different total number of experts $T$, specifically 80, 160, and 224 experts. We report the average score of 4 benchmarks, detailed results could be found in Table~\ref{table:totalnumber}, Appendix~ \ref{appen: detail_resutls_expert_number}. In addition to MoLA-$\triangledown$(2468), we introduce two variants that follow the MoLA-$\triangledown$ pattern, featuring a gradually increasing number of experts from lower layers to higher layers. For instance, MoLA-$\triangledown$(46810) allocates \textbf{4} experts to each layer for the first 8 layers, \textbf{6} experts to each layer for 9-16 layers, \textbf{8} experts to each layer for 17-24 layers, and \textbf{10} experts to each layer for the last 8 layers, which is denoted as \textbf{(46810)}. Figure~\ref{fig:expert_number} shows \ourmethod outperforms MoLA-$\triangledown$ across three configurations. Notably, \ourmethod with 80 experts surpasses MoLA-$\triangledown$(2468) with 160 experts, achieving comparable performance with 50\% fewer parameters. 

\section{Related work}
\subsection{Parameter-Efficient Tuning} 
As language models continue to expand in size, parameter-efficient tuning of LLMs has attracted significant interest due to its cost-effective approach for fine-tuning. Researchers introduce several PEFT approaches, such as LoRA~\citep{hu2021lora} and adapters~\citep{houlsby2019parameter}, aimed at enhancing the efficiency of fine-tuning large models. Among these, PEFT techniques based on low-rank adapters, called LoRA, have gained significant popularity and widespread adoption. These methods introduce two trainable low-rank matrices within each fully connected layer, resulting in substantial savings in training resources while maintaining performance. Building on these ideas, our approach integrates the MoE technique with LoRA adapters, employing layer-wise expert allocation to further enhance performance.

\subsection{LoRA-MoE Architecture} 
Recent research explores the integration of MoE~\citep{shazeer2017} and PEFT methods to boost performance in both single-task and multi-task scenarios \citep{li2024mixlora, huang2023lorahub, zhang2023composing, yang2024moralmoeaugmentedlora, dou-etal-2024-loramoe, feng-etal-2024-mixture, liu2023moelora, luo2024moelora}. For instance, \citet{liu2023moelora} employs LoRA and MoE for multi-task scenarios, particularly in healthcare. However, their methods necessitate the data type as input, limiting the model's applicability to other tasks. Similarly, \citet{dou2023art} propose LoRAMoE, a novel adapter architecture that integrates MoE and LoRA within the feed-forward layer of each Transformer block, addressing the issue of knowledge forgetting in LLMs during traditional supervised fine-tuning. Nonetheless, these approaches uniformly initialize the number of LoRA experts for each layer, resulting in redundancy among LoRA experts. \citet{gao2024higher} investigate redundancy in parameter-efficient MoE, initializing the number of experts with varying group-wise allocation and suggesting that higher layers require more LoRA experts. However, their allocation strategies are based on intuitive trial-and-error, lacking in-depth interpretability. Taking a step further, we introduce a fine-grained layer-wise expert allocation strategy by leveraging HT-SR theory to analyze layer training quality, improving the expert allocation in a theoretically principled manner.
\subsection{Heavy-Tailed Self-Regularization Theory.}
\citet{martin2019traditional, martin2020heavy, martin2021implicit, martin2021predicting} explore the HT properties observed in the ESD of weight matrices in neural networks. These HT structures provide insights into the underlying behavior and quality of models. Recent studies further demonstrate the utility of HT-SR Theory in various aspects of deep learning \citep{yang2023test, zhou2024temperature, lu2024alphapru, liu2024modelbalance,kothapalli2024crafting}. \citet{yang2023test} apply HT-SR principles to model selection, illustrating how the heavy-tailed characteristics can be leveraged to assess pre-trained NLP models without requiring training or testing data. Similarly, \citet{zhou2024temperature} extend this application to layer-wise adaptive training, showing that HT-SR can effectively guide the training process by assessing the quality of individual layers within a network. On the theory side, HT-SR theory has been substantially studied from various perspectives, including feature learning~\citep{wang2024spectral,kothapalli2024crafting} and overparameterization~\citep{hodgkinson2023interpolating}, which contribute to understanding the emergence of HT structures in the ESDs. Building on these insights, we utilize HT-SR theory to design an improved expert allocation method.
\section{Conclusion}
In this paper, we apply analytical methods from HT-SR theory to develop a fine-grained allocation strategy for determining the number of experts per layer in the LoRA-MoE architecture. Extensive empirical results show that \ourmethod offers a simple yet effective approach to layer-wise expert allocation. Analysis of three widely used transformer-based language models reveals that well-trained layers require fewer LoRA experts. Our theoretically grounded method provides a scalable solution for various models, further reducing redundancy within the LoRA-MoE architecture. In future work, we aim to integrate \ourmethod with other PEFT methods and explore its application across different model architectures and domains.

\section{Limitations}
\ourmethod demonstrates both effectiveness and scalability as a model-specific method for determining the number of LoRA experts. However, there are some potential limitations to consider. First, the performance of \ourmethod could be varied for different tasks, which could increase the uncertainty in the performance of this method. Additionally, the optimal total number of experts is determined by the experimental results. Overall, we will continue to work on this problem to address these limitations and develop more effective and robust allocation methods for different tasks.

\section*{Ethics Statement}
We have not identified any ethical concerns directly related to this study.

\section*{Acknowledgment}
This research was partially funded by a Google Research Award.


\bibliography{emnlp2023.bib}
\bibliographystyle{acl_natbib.bst}

\clearpage
\appendix

\section{Definition of other shape metrics from HT-SR theory}
\label{appen:def_metric}
\begin{itemize}

    \item (\texttt{Stable\_Rank\xspace}) The \texttt{Stable\_Rank\xspace} metric provides a norm-adjusted assessment of the scale of the empirical spectral density (ESD). Prior research \cite{martin2021predicting} has shown that \texttt{Stable\_Rank\xspace} correlates with the \texttt{PL\_Alpha\xspace} metric. For a given weight matrix $\mathbf{W}$, it is calculated as follows:
    \begin{equation}
        \texttt{Stable\_Rank\xspace} = \frac{\|\mathbf{W}\|_F^2}{\|\mathbf{W}\|_2^2},
    \end{equation}
    where $\|\mathbf{W}\|_F$ denotes the Frobenius norm and $\|\mathbf{W}\|_2$ denotes the spectral norm.
    
    \item (\texttt{Alpha\_Hat\xspace}) The \texttt{Alpha\_Hat\xspace} metric, introduced in \cite{martin2021predicting}, has been demonstrated to be effective at predicting model generalization. It represents a modified form of the Power Law (PL) exponent $\alpha$ (\texttt{PL\_Alpha\xspace}), scaled by the logarithm of the largest eigenvalue $\lambda^{\text{max}}$ of the weight matrix's spectral norm (\texttt{log\_spectral\_norm\xspace}):
    \begin{equation}
        \texttt{Alpha\_Hat} = \alpha \log \lambda^{\text{max}}.
    \end{equation}
    
\end{itemize}

\section{Dataset}
\label{appen:dataset}
\ourmethod is studied on ten standard datasets from three categories:
\subsection{Language Understanding}
\paragraph{Microsoft's Research Paraphrase Corpus (MRPC).} This dataset has 5,801 sentence pairs from news articles, labeled to indicate paraphrases. It includes 4,076 pairs for training and 1,725 for testing, with the task of classifying paraphrase pairs.
    
\paragraph{Recognizing Textual Entailment (RTE).} Derived from annual textual entailment challenges (RTE1, RTE2, RTE3, RTE5), this dataset features sentences from news and Wikipedia, classified as entailment or not. It comprises 2,490 training and 277 validation samples.
\paragraph{Corpus of Linguistic Acceptability (COLA).} This dataset contains English sentences annotated for grammatical acceptability, sourced from linguistic theory texts. It includes 8,551 training and 1,043 validation samples.

\subsection{Commonsense Reasoning}
\paragraph{ScienceQA.} This dataset includes 21,208 multiple-choice questions from elementary and high school science curricula. For text-only samples, there are 6,508 training and 2,224 test samples, covering natural science, language science, and social science, requiring commonsense knowledge for answers.
\paragraph{CommonsenseQA.} A dataset for commonsense reasoning with 9,740 training samples and 1,221 validation samples, created by Amazon Mechanical Turk workers. It demands various types of commonsense knowledge to determine the correct answers.
\paragraph{OpenbookQA.} Comprising 5,957 elementary-level science questions, this dataset tests understanding of core science facts and their application to new scenarios. It includes 4,957 training, 500 validation, and 500 test samples.

\subsection{Arithmetic Reasoning}
Table~\ref{appen:math_datasets} presents the statistics of four Arithmetic Reasoning benchmarks: (1)AddSub \cite{addsub}, (2) MultiArith \cite{multiarith}, (3) SVAMP \citep{patel2021svamp}, (4) GSM8K \citep{cobbe2021gsm8k}. 
\begin{table}[!ht]

  \centering
  
  \begin{adjustbox}{width=0.70\columnwidth}
  \begin{tabular}{lcc}
    \toprule
    \textbf{Dataset} & \textbf{\# of Samples} & \textbf{Avg. Words} \\
    \midrule
    AddSub & 395 & 31.5 \\
    MultiArith & 600 & 31.8 \\  
    SVAMP & 1000 & 31.8 \\
    GSM8K & 1319 & 46.9 \\
    \bottomrule
  \end{tabular}
  
  \end{adjustbox}
  \caption{Statistics of Arithmetic Reasoning datasets.}
  \label{appen:math_datasets}
\end{table}

\begin{figure*}[!ht]
\begin{center}
\centerline{\includegraphics[width=2\columnwidth]{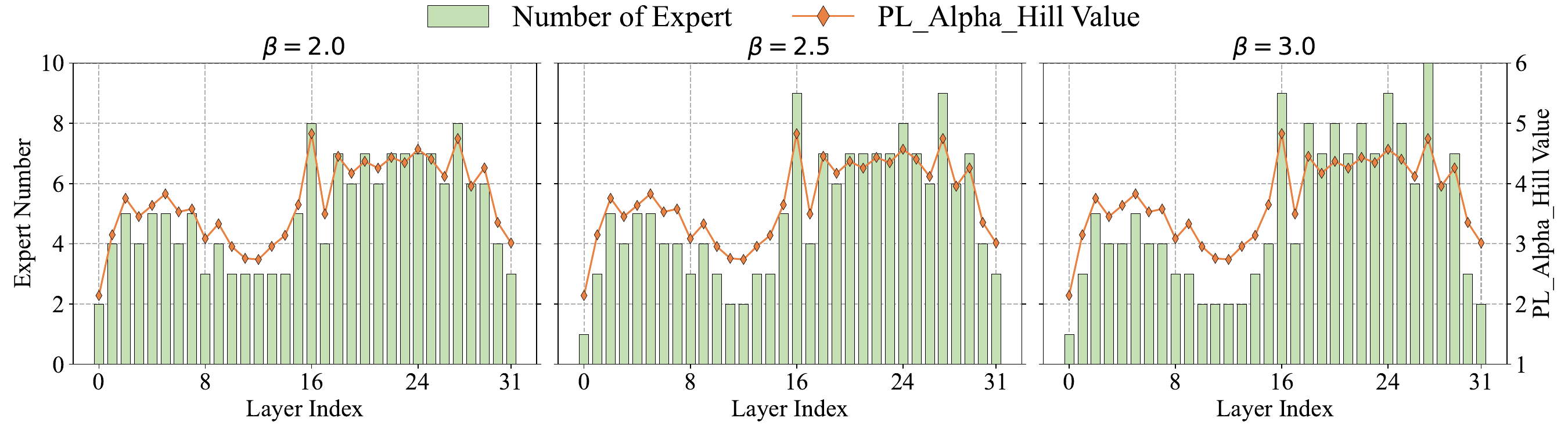}}
\caption{The expert allocations for Mistral-7b under different parameter $\beta$.}
\label{fig4:beta} 
\vspace{-5mm}
\end{center}
\end{figure*}
\begin{table*}[!ht]
    \centering    
    \small
    
        \begin{tabular}    {c | l | c  c  c  c }
        \toprule        
        Expert Amount & Strategy &  GSM8K & SVAMP & AddSub & MultiArith \\
        \midrule
        \multirow{2}{*}{\textbf{80}}&\textbf{MoLA-$\triangledown$(1234)} & 68.23 & 73.90 & 56.70 & 95.33 \\

        &\textbf{\ourmethod} & \textbf{69.44} & \textbf{75.60} & \textbf{57.72} & \textbf{97.66} \\
        \midrule
        \multirow{2}{*}{\textbf{160}}&\textbf{MoLA-$\triangledown$ (2468)} & 67.20 & 76.50 & 59.24 & 97.00 \\

        &\textbf{\ourmethod} & \textbf{68.30} & \textbf{77.20} & \textbf{59.49} & \textbf{97.33} \\
        \midrule
        \multirow{2}{*}{\textbf{224}}&\textbf{MoLA-$\triangledown$(46810)} & 68.61 & 73.60 & \textbf{58.98} & 96.83 \\

        &\textbf{\ourmethod} & \textbf{69.06} & \textbf{75.30} & 58.22 & \textbf{97.50} \\

        \bottomrule    
        \end{tabular}
        \caption{Comparison between \ourmethod and MoLA-$\triangledown$ across three configurations with varying total number of experts $T$, specifically 80, 160, and 224 experts. }
    \label{table:totalnumber}
\end{table*}

\begin{table}[!ht]
    \centering
    \small
    \begin{tabular}{l c c c}
    \toprule
    Strategy & MRPC & COLA & RTE \\
    \midrule
    \textbf{MoLA-$\triangledown$} & 83.48 & \textbf{86.87} & 86.28 \\
    \textbf{\stablerank} & \textbf{84.34} & 84.56 & 86.64 \\
    \textbf{\alphahat} & 84.11 & 86.86 & 84.83 \\
    \textbf{\ourmethod} & 84.23 & 86.67 & \textbf{87.36} \\
    \bottomrule
    \end{tabular}
    \caption{Comparison of shape metrics on GLUE tasks.}
    \label{table:results1}
\end{table}

\begin{table}[!ht]
    \centering
    \small
    \begin{tabular}{l  c c c }
    \toprule
    Strategy & S.QA & C.QA & O.QA  \\
    \midrule
    \textbf{MoLA-$\triangledown$} & 92.36 & \textbf{78.95} & 79.60  \\
    \textbf{\stablerank} & 92.13 & 77.81 & 77.80 \\
    \textbf{\alphahat} & 91.86 & 78.13 & 79.80  \\
    \textbf{\ourmethod} & \textbf{92.71} & 78.37 & \textbf{80.80}  \\
    \bottomrule
    \end{tabular}
    \caption{Comparison of shape metrics on QA tasks (S.QA, C.QA, O.QA denote ScienceQA, CommonsenseQA, and OpenbookQA respectively).}
    \label{table:results2}
\end{table}

\section{Implementation}
\label{appen: implement}
The direct fine-tuning setting aligns with \citet{gao2024higher}, we do a grid search on the number of training epochs, including 10, 15, and 20 epochs for downstream task fine-tuning. The cutoff length is set to 256 and the batch size is 128. For the second instruction-tuning$\rightarrow$zero-shot tasks evaluation, we conduct instruction-tuning on the MetaMathQA dataset \cite{yu2023metamath} for 1 epoch with cutoff length set to 512. We conduct a small hyperparameter sweep within the range of $\beta \in [2.0, 2.5, 3.0]$, where $\beta$ regulates the standard deviation of experts allocation, as depicted in Figure~\ref{fig4:beta}. For all experiments, we use AdamW~\citep{loshchilov2017decoupled} as the optimizer with a learning rate of 3e-4. The rank of each LoRA expert is 8 and we adopt Top-2 for the router. LoRA alpha is set to 16 and LoRA dropout is 0.05, following the default LoRA settings~\cite{hu2021lora}. We apply LoRA experts to four weight matrices in the self-attention module ($\mathbf{W_q}$, $\mathbf{W_k}$, $\mathbf{W_v}$, $\mathbf{W_o}$) and three weight matrices in the MLP module ($\mathbf{W_{gate}}$, $\mathbf{W_{down}}$, $\mathbf{W_{up}}$). All experiments are conducted with three RTX A6000-48G GPUs.

\begin{table}[!ht]
    \centering
    \small
    \resizebox{\columnwidth}{!}{ 
    \begin{tabular}{l c c c c}
    \toprule
    Shape Metrics & GSM8K & SVAMP & AddSub & MultiArith \\
    \midrule
    \textbf{MoLA-$\triangledown$} & 50.11 & 56.40 & \textbf{48.86} & 87.66 \\
    \textbf{\stablerank} & 48.22 & 55.20 & 48.61 & 88.50 \\
    \textbf{\alphahat} & 49.81 & 56.70 & 46.08 & 88.50 \\
    \textbf{\ourmethod} & \textbf{50.41} & \textbf{57.00} & 48.60 & \textbf{91.33} \\
    \bottomrule
    \end{tabular}
    }
    \caption{Comparison of shape metrics on arithmetic tasks.}
    \label{table:shape_metrics_zeroshot}
\end{table}

\section{Complementary Results}
\label{appen: detail_results}
In this section, we provide detailed results in Section~\ref{sec4.3: shapemetric} and Section~\ref{sec4.4: expertnumber}.

\subsection{Detailed Results of Different Shape Metric}
\label{appen: detail_resutls_shapem}
Table~\ref{table:results1} and~\ref{table:results2} present the results of several shape metrics in HT-SR theory when directly fine-tuned on GLUE and QA tasks. Table~\ref{table:shape_metrics_zeroshot} shows the results of zero-shot setting on math tasks. We report the average score for both settings in Figure~\ref{fig: shapemetric}.

\subsection{Detailed Results of Varying Number of Experts}
\label{appen: detail_resutls_expert_number}
In Table~\ref{table:totalnumber}, we show the results of \ourmethod and MoLA-$\triangledown$ across three configurations with varying total number of experts $T$, specifically 80, 160, and 224 experts.

\end{document}